# Vision-based Structural Inspection using Multiscale Deep Convolutional Neural Networks

V. Hoskere[1], Y. Narazaki[2], T.A. Hoang[3], B.F. Spencer Jr.[4]

1. *Ph.D. student, Department of Civil and Environmental Engineering, University of Illinois at Urbana-Champaign, U.S.A. E-mail:* hoskere2@illinois.edu
2. *Ph.D. student, Department of Civil and Environmental Engineering, University of Illinois at Urbana-Champaign, U.S.A. E-mail:* narazak2@illinois.edu
3. *Ph.D. student, Department of Civil and Environmental Engineering, University of Illinois at Urbana-Champaign, U.S.A. E-mail:* tuhoang2@illinois.edu
4. *Professor, Department of Civil and Environmental Engineering, University of Illinois at Urbana-Champaign, U.S.A. E-mail:* bfs@illinois.edu

**ABSTRACT**
Current methods of practice for inspection of civil infrastructure typically involve visual assessments conducted manually by trained inspectors. For post-earthquake structural inspections, the number of structures to be inspected often far exceeds the capability of the available inspectors. The labor intensive and time consuming natures of manual inspection have engendered research into development of algorithms for automated damage identification using computer vision techniques. In this paper, a novel damage localization and classification technique based on a state of the art computer vision algorithm is presented to address several key limitations of current computer vision techniques. The proposed algorithm carries out a pixel-wise classification of each image at multiple scales using a deep convolutional neural network and can recognize 6 different types of damage. The resulting output is a segmented image where the portion of the image representing damage is outlined and classified as one of the trained damage categories. The proposed method is evaluated in terms of pixel accuracy and the application of the method to real world images is shown.
.
**KEYWORDS:** *Damage detection, Damage localization, Computer vision, Machine learning, Multi-scale deep convolutional neural networks.*

## 1. INTRODUCTION

Structural inspections are important tools for assessment of the current condition of infrastructure to ensure their safety and serviceability. Government agencies spend vast sums of money conducting periodic inspections on critical infrastructure like bridges and dams. In a post disaster scenario, there is often a shortage of funds, labor and time to carry out inspections of all affected buildings and transportation infrastructure. Despite recent advances in technology, manual visual inspection remains the main method of condition assessment of civil infrastructure in the United States [1]. Such manual inspections have obvious drawbacks: they are time consuming, labor intensive, subjective and often unsafe. Catastrophic accidents like the I-35W bridge collapse in Minneapolis reveal that manual inspections can also be unreliable, despite following best practices [2] as it is easy to miss important details. A natural step forward is to reduce to amount of effort required to carry out such inspections using automation.

The application of computer vision techniques holds the promise of automating the structural inspection process by providing artificial intelligence that can be deployed on self-navigating robots such as UAVs or ground robots. While several issues have to be addressed before such automated inspections become a reality, the aspect germane to the current investigation is the ability to automatically assess the presence of a variety of structural damage types without human intervention. A number of vision based methods have been created for the purpose of identifying defects in civil infrastructure. Many such methods can be characterized as utilizing a hand crafted filtering technique to identify some intensity variation. Initial methods focused on identifying concrete cracks, rely on some threshold parameters sensitive to ambient lighting [3], [4], [5]. In [6] and [7] depth and 3D information were employed to obtain improved results. Recently, computer vision techniques have also been used to identify other concrete defects like spalling. In [8], a combination of segmentation, template matching and morphological pre-processing were employed, both for spall detection and assessment on concrete columns. A novel orthogonal transformation approach combined with a Bridge Condition Index was used in [9] to quantify degradation and subsequently map to condition ratings. The authors were able to achieve a reasonable accuracy of 85% but were not able to address situations where both cracks and spalls were present. Computer vision techniques have been used for identification of defects like corrosion and fatigue cracks in steel structures. In [10] the authors employ a support vector machine to identify rust on steel bridges and in [11], corrosion detection in navigational vessels was carried out using a combination of image based weak learners together with AdaBoost. Research about fatigue crack detection in civil infrastructure has been fairly limited. In [12] the authors manually created defects on a steel beam to give the appearance of fatigue cracks. A combination of region localization by object detection and filtering techniques were then used to identify the created fatigue crack like defects. The authors assumed that fatigue cracks generally developed around bolt holes, but this is not always true for a variety of steel structures including, for example, navigational infrastructure like miter gates. There has been extensive work on crack detection in asphalt pavements. In [13] a local binary patterns (LBP) based algorithm was used to identify cracks, in [14] a Gabor filtering technique was employed and in [15] filter based features are used together with a classifier to identify cracks for subway tunnel safety.

The studies and techniques described thus far can be categorized as either having used unsupervised techniques or relying on a combination of hand crafted features together with a classifier. In essence however, the application of such techniques in an automated structural inspection environment is limited because these techniques do not employ contextual information available in the regions around where the defect is present. Real world situations vary extensively and it is thus difficult to hand craft a general algorithm that can be successful in varying environments. The success of convolutional neural networks(CNNs) and deep learning algorithms(DLAs) in computer vision [16] [17] [18] in recent years has had a significant impact in a number of fields such as general image classification, autonomous transportation systems, and medical imaging. Recently, there has been work on the application of CNNs and DLAs to problems of damage identification and automated structural inspection. [19] utilized a CNN for the extraction of important regions of interest in highway truss structures to ease the inspection process. CNNs have also recently been employed for the application of crack detection in asphalt pavements [20] and Deep CNNs (DCNNs) were tested for concrete crack identification with very high accuracy in both cases. R-CNNs have been used for spall detection in a post disaster scenario [21], although, the results (about 59.39% true positive accuracy) leave much scope for improvement.

A major drawback of the methods that have been proposed thus far is that they are only applicable to a single type of damage. One of the many advantages of deep learning methods however is the ability of deep networks to learn general representations of identifiable characteristics in images. For example, DCNNs have been successful in classification problems with over 1000 classes [16]. In our present work, we propose the application of a deep learning method for general damage detection, i.e., detection of multiple types of visible damage using a single algorithm. Needless to say, real structures often have different members made of different building materials and are also susceptible to a myriad of surface defects. Thus, general or multiclass damage detection is an important problem to be addressed [9] [22] before a truly automated structural inspection system can be successful.

In this study, we use a multiscale pixel-wise deep convolutional neural network (MPDCNN) for the purpose of general damage localization and classification of six different types of damage, namely - concrete cracks, concrete spalling, exposed reinforcement bars, steel corrosion, steel fracture and fatigue cracks and asphalt cracks. We address several critical drawbacks of previous works of automated vision based damage detection. We include a larger context of the image in the algorithms by employing a multi-scale CNN. Most unsupervised methods do not use contextual information and the CNN based methods proposed thus far use only local context. By employing a multiscale CNN, a larger context of pixels around the damage is included allowing the network to better learn the context [23] in which damage may occur. Furthermore, since the same filters are used at multiple scales, the trained algorithm is also approximately scale invariant. Since each pixel is treated as a data point, unlike ordinary CNNs that require large number of images of the order of $\sim 10^5$, the proposed MPDCNN can be trained with significantly fewer images ($\sim 10^3$). The presence of multiple scales also helps to reduce overfitting. In order to test the proposed algorithm, we use a dataset mined from various sources consisting of images of a variety of structures including buildings, navigational infrastructure and coastal infrastructure such as dams and lock gates, bridges and pavements.

## 2. AUTOMATED VISION BASED DAMAGE LOCALIZATION AND CLASSIFICATION

We propose a new method for localization and classification for six different types of damage found in civil infrastructure. The pipeline consists of a two parallel steps, both utilizing MPDCNNs. The first network, hereafter referred to as the damage classifier, independently classifies each pixel into one of seven classes, namely no-damage, concrete cracks, concrete spalls, exposed reinforcement, steel corrosion, steel fatigue cracks and asphalt cracks. The next step involves a parallel network that serves as a "damage segmenter" which simply differentiates between pixels that represent damage and pixels that do not, i.e., the segmenter performs a binary pixel-wise classification. The purpose of this parallel damage segmenter is to reduce false positives that occur in the classification and results show that the inclusion of the second network also provides finer segmentation. Finally, information the labels produced by the segmenter and classifier are combined using simple rules.

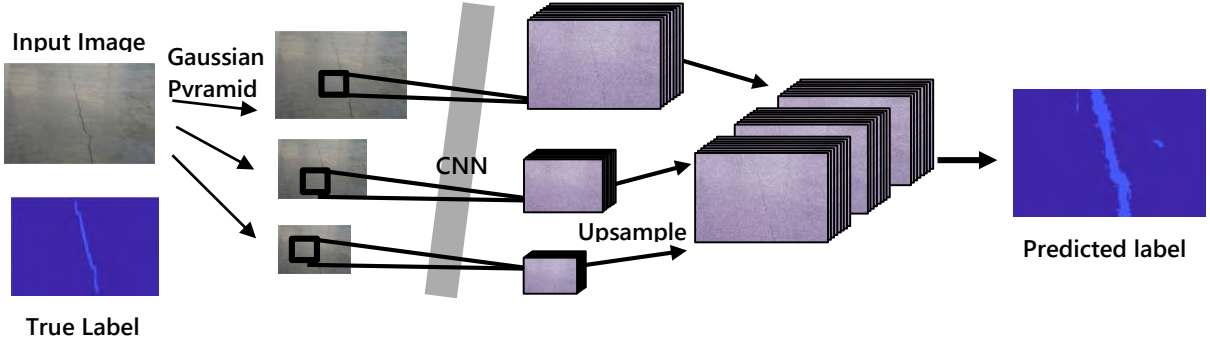

Figure 1. Pipeline for pixel-wise training using multiscale convolutional neural networks

## 2.1 Multi-scale Pixel-wise Deep Convolutional Neural Networks

As mentioned earlier, we utilize MPDCNNs for both the damage segmenter and damage classifier. Multi-scale convolutional networks, first proposed in [23], learn to generate meaningful feature vectors for regions of multiple sizes all centred around the same pixel. In natural scenes, features extracted are expected to be hierarchical. Hence, a logical approach is to use multi-layer convolutional neural network in which lower level layers learn lower features on the hierarchical map (e.g.: pixel, edges, motif, parts, objects). For maximal efficiency and scale invariance, each scale utilizes the same network and thus the trained network. The features learnt are representations that capture texture, shape and contextual information. The MPDCNN architecture utilized in this study differs from that used in [23] in two salient ways. Firstly, we utilize deep network architectures, namely a 23 layer ResNet [16] for the damage segmenter, and a modified form of the VGG19 [17] network for the damage classifier. Further, we utilize an RGB Gaussian pyramid to generate data at multiple scales, as opposed to a HSV based Laplacian Pyramid.

A multiscale Gaussian pyramid, $X_s$ with three layers is constructed by resizing the original image, $I$ by a factor of 2 and a factor of 4. The generated pyramid is then fed into a deep convolutional neural network (DCNN), $f_s$ consisting of several hidden layers. Each hidden layer in the DCNN has three main parts: (i) a linear filter bank layer (Eq. (1) (ii) a pixel-wise non-linear transformation (2) and (iii) a feature pooling/ down-sampling layer (3).

$$H_{conv\{l\}} = W_{conv\{l\}} H_{\{l\}} + b_{\{l\}} \qquad (1)$$
$$H_{act\{l\}} = f_a(H_{conv\{i\}}) \qquad (2)$$
$$H_{\{l\}} = \text{maxpool}_{n \times n}(H_{act\{l\}}) \qquad (3)$$

$W_{conv}$ is a Toeplitz weight matrix representing a convolution, $b$ is an additive bias, $f_a$ is a nonlinear activation function, $n$ denotes the stride of the maxpooling operation, $l$ denotes the $l^{th}$ layer and $H_0 = X_s$. Each scale is passed through the convolution network and outputs feature maps that are the same size as the scaled images. These feature maps are upsampled and concatenated to form a multiscale feature representation $F_i$ (4).

$$H_{\{L+i=1\}} = [H_{L_{s=1}}, u_2(H_{L_{s=2}}), u_4(H_{L_{s=4}})] \qquad (4)$$

$H_{\{L+i\}}$ is a 3d matrix whose first two dimensions correspond to the size of the input image and last dimension correspond to the three times the number of features maps output by the DCNN, $L$ corresponds to the number of layers in the network excluding the fully connected layers (5) and $i$ corresponds to the $i^{th}$ fully connected layer.

$$H_{\{L+i+1\}} = f_a(W_{fc\{i\}} H_{\{L+i\}} + b_{fc\{i\}}) \qquad (5)$$

Again, $W_{fc}$ represents a fully connected weight matrix and $b_{fc\{i\}}$ is a bias. The size of the final weight matrix is such that the final fully connected layer has as many dimensions as the number outputs desired, i.e., 7 for the damage classifier and 2 for the damage segmenter. The final output layer is then scaled to (0, 1) using a softmax squashing function (6).

$$\hat{y}_{p,c} = \frac{e^{H_{\{L+I\}}(p,c)}}{\sum_{c \in C} e^{H_{\{L+I\}}(p,c)}} \tag{6}$$

$\hat{y}_{p,c}$ denotes the output of the MPDCNN in the form of a probability per class for the $p^{th}$ pixel and $c^{th}$ class out of a total of $C$ classes. The actual class assigned to a pixel will then simply be $\hat{y}_p = argmax(\hat{y}_{p,c})$.

**2.2 Training considerations**

The network parameters, namely $W$ and $b$, are trained by minimizing the cross-entropy loss function between the predicted softmax probabilities and the corresponding one-hot labels with an L2-regularization weight decay.

$$W, b = argmin\left(-\sum_{p,c} y_{p,c} \ln(\hat{y}_{p,c}(W,b)) + \lambda \sum_{l}^{L+I} ||W_l||^2\right) \tag{7}$$

$y_{p,c}$ corresponds to pixel-wise one hot labels, i.e., $y_{p,c=k} = 1$, if $label_p = k \in C$, and 0 otherwise. To the operations described in section 2.1, certain augmentations are incorporated in order to increase the efficacy and efficiency of training and prevent issues such as overfitting. A complication of training deep networks is that the distribution of each layer's inputs changes during training, as the parameters of the previous layers change. When using saturating activations such as a softmax, distribution shifts may slow down training significantly since the activation may cause the output of a layer to fall in a saturated region. In order to solve this problem, we apply batch normalization proposed in [24], where each feature dimension is shifted by a weighted mean and standard deviation that is learnt during training. We also apply dropout [25], a technique meant to mitigate overfitting by randomly sampling selected layers with a certain probability instead of propagating all dimensions of the layer. To balance the frequencies of different classes in our data set and prioritize all classes equally, we apply median class balancing [26] which reweights each class in the cross entropy loss. Next, in section 2.3 we describe the network architectures employed in this study

| VGG19_reduced | | \multicolumn{5}{c}{ResNet23} | | | |
|---|---|---|---|---|---|---|---|
| Name | Filt. Size | Name | Filt. Size | ResNet connect. | Name | Filt. Size | ResNet connect. |
| Conv0 | 3x3x64 | Conv0 | 7x7x32 | | Conv18 | 3x3x128 | Conv16 |
| Conv1 | 3x3x64 | Conv1 | 7x7x32 | | Conv19 | 3x3x128 | |
| Maxpool0 | 2x2 | Conv2 | 7x7x32 | Conv0 | Conv20 | 3x3x128 | Conv18 |
| Conv2 | 3x3x128 | Maxpool0 | 2x2 | | FCL0 | 1024 | |
| Conv3 | 3x3x128 | Conv3 | 3x3x64 | | FCL1 | 7 or 2 | |
| Maxpool1 | 2x2 | Conv4 | 3x3x64 | Maxpool0 | Batch size | 5 | |
| Conv4 | 3x3x256 | Conv5 | 3x3x64 | | Wt. decay | 0.0001 | |
| Conv5 | 3x3x256 | Conv6 | 3x3x64 | Conv4 | Dropout | 85% (FCL1) | |
| Conv6 | 3x3x256 | Conv7 | 3x3x64 | | #param | 2143618 (segmenter) | |
| Conv7 | 3x3x256 | Conv8 | 3x3x64 | Conv6 | | 2148743 (classifier) | |
| FCL0 | 1024 | Conv9 | 3x3x128 | | | | |
| Name | Filt. Size | Conv10 | 3x3x128 | Conv8 | | | |
| FCL1 | 1024 | Maxpool1 | 2x2 | | | | |
| FCL2 | 256 | Conv11 | 3x3x128 | | | | |
| FCL3 | 7 or 2 | Conv12 | 3x3x128 | Maxpool1 | | | |
| Batch size | 5 | Conv13 | 3x3x128 | | | | |
| Wt. decay | 0.0005 | Conv14 | 3x3x128 | Conv12 | | | |
| Dropout(FCL0, FCL1) | 0.85 | Conv15 | 3x3x128 | | | | |
| #param | 4421824 (segmenter) | Conv16 | 3x3x128 | Conv14 | | | |
| | 4423104 (classifier) | Conv17 | 3x3x128 | | | | |

Table 1. Network architectures tested for the purposes of automated structural inspections. VGG19_reduced is derived from the VGG19 network proposed in [17] and the RestNet23 architecture is a scaled down version of ResNet45 proposed in [16].

## 2.3 Network architectures

We employ different network architectures for the two parallel networks in this study. The network architecture proposed in the first work on multiscale CNNs [23] was shallow by current standards. Several more successful network architectures have been proposed since, two notable architectures being the VGG networks [20] and ResNet45 [16]. Residual networks such as ResNet45 provide one way to ensure that the addition of layers enforces the network to learn new concepts by supplying hidden layers a shortcut link from earlier layers in the network. The main idea is to prevent additional deeper layers from becoming redundant while also mitigating the effect of vanishing gradient due to increased number of layers. We adopt scaled down versions of both these architectures in our study, to gain insight into which networks perform better for the tasks of segmentation and classification.

The details of the networks employed are shown in Table 1. VGG19_reduced is derived from the VGG19 network proposed in [17], with the main difference in the architecture being that the only the first 8 convolutional layers were used and the sizes of the fully connected layers have been reduced. Similarly, the ResNet23 architecture used was derived from the ResNet45 architecture proposed in [16] and is similar to the architecture employed in [27]. With the hope of better capturing local damage features, we increase the size of the first three layers to convolutions of size 7x7 but with shallower depths. The depths of the filters are increased after every maxpool operation and the residual connections are made between alternate layers. Thus we employ networks based on proven, successful architectures and with modifications aimed at adapting them to the more specific problem of automated structural inspections.

## 3. GENERATING A DATABASE FOR STRUCTURAL DAMAGE

There currently exists no publicly available dataset of labelled images showing structural damage. Thus, a new database had to be assembled from scratch to test the proposed method. To demonstrate the generality of the proposed method, it was important to include images from a variety of different types civil infrastructure. To this end, different damaged specimens were photographed by the authors, and images of full structures available in the public domain on the internet were included. The other sources included: datacentrehub.org [28], bridgehunter.com [29], images available on the website of the US Army Corps of Engineers [30] as well as images acquired from google image searches. Overall, the assembled database includes images of reinforced concrete buildings, steel bridges, concrete bridges, asphalt pavements, hydraulic structures, inland navigation infrastructure, concrete pavements and damage laboratory specimen. The database includes a total of 339 images of over 250 different structures of varying sizes that have been cut into 1695 images of a uniform size 600x600. The details of these images is given in Table 2. All of the images were hand labelled by the authors using a Matlab GUI created by the authors for pixel-wise labelling of images.

Table 2. Details of dataset of damaged structures generated for this study.
(Images have been acquired from various sources and hand labelled by the authors)

| Main type of damage in image (each image may also include other types of damage or no damage) | Number of images |
|---|---|
| Concrete Spalling, Exposed Reinforcement | 324 |
| Fatigue Cracks | 216 |
| Concrete Cracks | 341 |
| Steel Corrosion | 379 |
| Asphalt Cracks | 435 |
| Total | 1695 |

## 4. RESULTS OF AUTOMATED DAMAGE ASSESMENT

The network architectures described in section 2 were implemented on the generated database to test the efficacy of their performance for the purpose of automated damage assessment. In order to allow for quicker training, the 1695 labelled images of size 600x600x3 were resized to 288x288x3. To artificially increase the amount of data, a data augmentation strategy was implemented. We followed the suggestions made in [23] with random resizing with factors uniformly distributed between 0.75 and 1.25, random rotations between $\pm 15°$, random flipping and white noise with standard deviation of 2. 80% of the images were used for training purposes and the remaining 20% were set aside for testing purposes, to validate the network.

## 4.1 Damage Classification and Segmentation

The labelled data was fed into each network in batches of 5 images at a time. Both networks proposed in section 2 were tested for each the damage segmenter and the damage classifier. The training was carried out on a Windows PC with an i7 7700 2.8GHz processor, NVIDIA GTX 1070 8Gb graphics card and 16GB RAM. The learning rate used for both the networks was $10^{-3}$ for the first 70 iterations, $10^{-4}$ for the next 50 cycles, $10^{-5}$ for the next 25 cycles and $10^{-6}$ for the final 15 cycles. It was found that the ResNet23 architecture performed better for segmentation with a test accuracy of 88.8% compared to an accuracy of 87.2% from training with VGG19_reduced. On the contrary, the VGG19 served as a better classifier with an accuracy of 71.4% vs 68.3% from training with ResNet23. The results from the segmenter and classifier were combined using appropriate softmax thresholds that was manually chosen for each of the classes (cracks:0.1, spalls:0.4, exposed reinforcement: 0.1, corrosion:0.5, fatigue cracks:0.1, asphalt cracks: 0.5). The resulting accuracy was 86.7%, much improved from the 71.4% of only the damage classifier, while it can also be seen that there is a slight reduction in damage class accuracies for spalling, asphalt cracks and corrosion but each class still has an accuracy of over 80%. These results are summarized in Figure 2. Sample results of the automated inspection is shown in Figure 3.

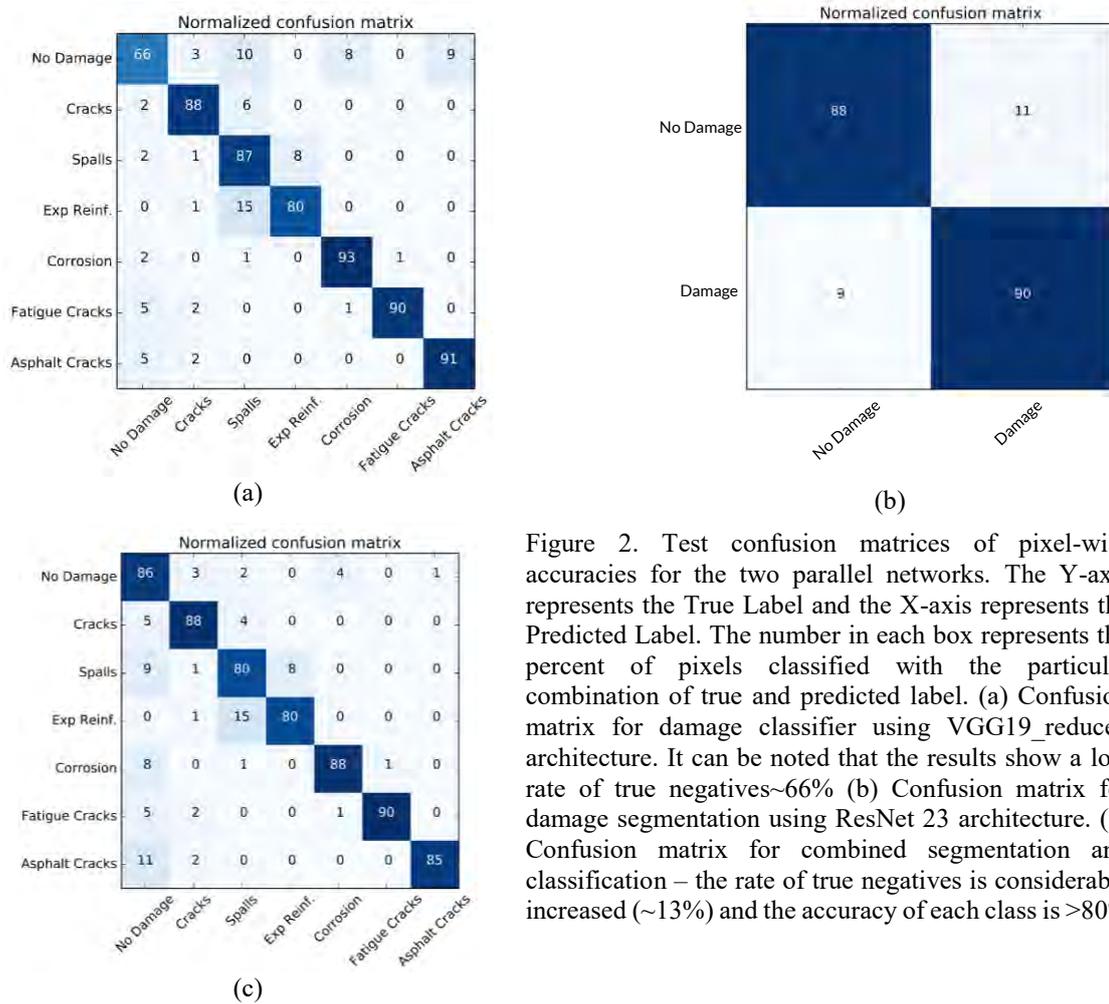

Figure 2. Test confusion matrices of pixel-wise accuracies for the two parallel networks. The Y-axis represents the True Label and the X-axis represents the Predicted Label. The number in each box represents the percent of pixels classified with the particular combination of true and predicted label. (a) Confusion matrix for damage classifier using VGG19_reduced architecture. It can be noted that the results show a low rate of true negatives~66% (b) Confusion matrix for damage segmentation using ResNet 23 architecture. (c) Confusion matrix for combined segmentation and classification – the rate of true negatives is considerable increased (~13%) and the accuracy of each class is >80%

## 5. CONCLUSIONS

We present a new method for automated vision based structural inspection using multiscale pixel-wise deep convolutional neural networks. To the authors' best knowledge, this is the first published work where a general purpose damage detection strategy – capable of handling multiple types of damage classes simultaneously, for a wide range of civil infrastructure, and where damage is delineated pixel-wise – has been proposed. The proposed method was tested on a dataset of 1695 images with six different types of damage assembled from over 250 different structures. Two parallel networks were utilized, one that served as a damage segmenter and another that served as a damage classifier. Two different architectures were tested for each of the two networks to select the architecture with best performance. The classifier was able to perform an accurate classification between different types of damage and the segmenter was able to reduce the rate of false positives by delineating the portions of the image that represented damage. The combined results showed a high accuracy of 86.7% across 7 classes. While much work is still required in the improvement in robustness of such algorithms, in the future, the authors envisage a scenario where such networks can be deployed on robots such as unmanned aerial vehicles to carry out structural inspections in an automated fashion for time and cost sensitive applications or when it is unsafe for humans.

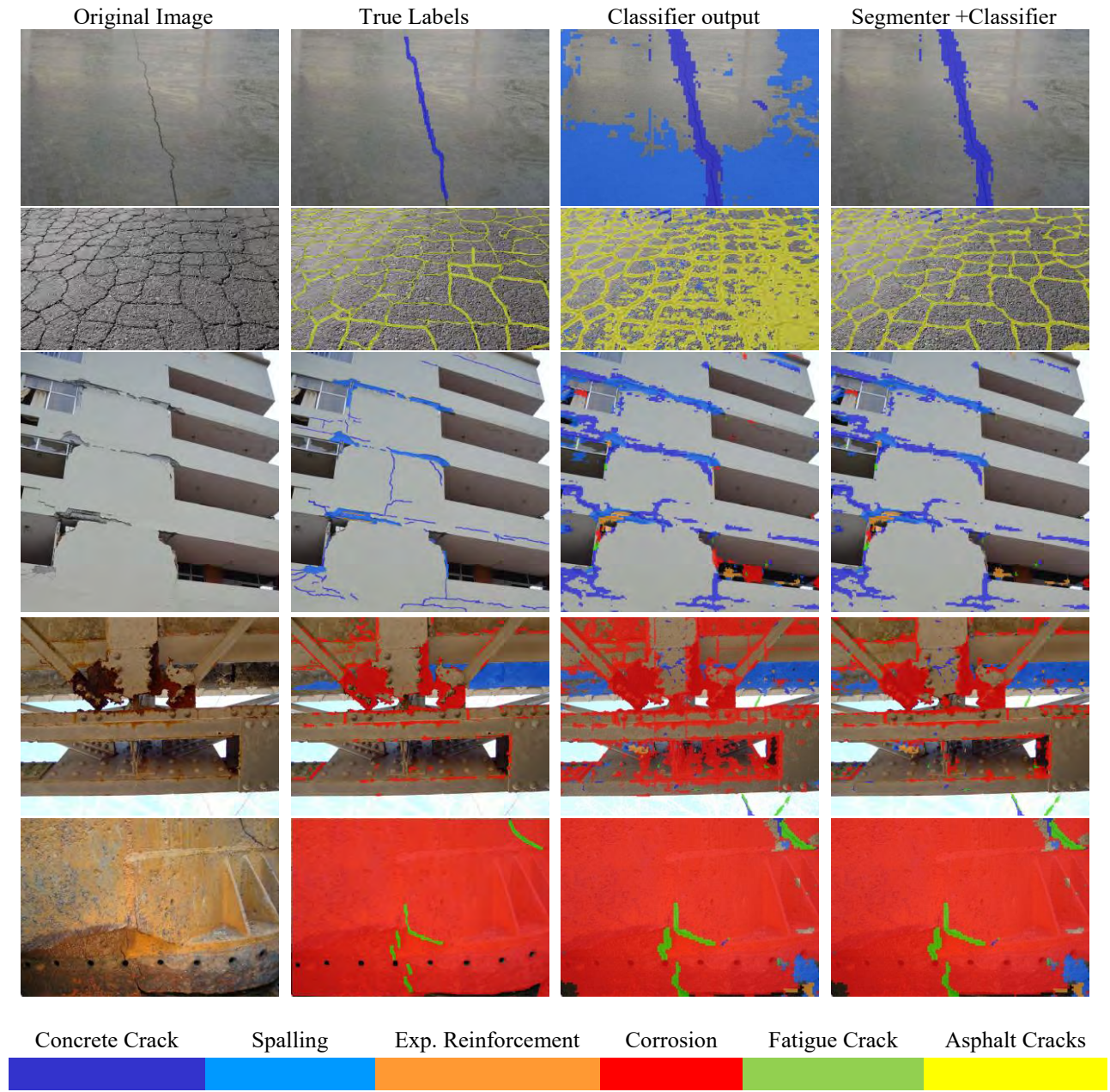

Figure 3. Sample results of automated structural inspection using the proposed methods and a comparison of outputs with and without the damage segmenter. Each color represents a different damage type as shown.